\documentclass[a4paper,fleqn]{cas-dc}

\usepackage[authoryear]{natbib}

\usepackage{siunitx}
\usepackage{amsmath,amssymb,amsfonts}
\usepackage{mathtools}
\usepackage{pdflscape}
\usepackage{multirow}
\usepackage{threeparttable}
\usepackage{flushend}
\usepackage[subnum,fleqn]{cases}

\graphicspath{{./Figures/}{./photos/}}
\DeclareGraphicsExtensions{.pdf,.png}

\def\tsc#1{\csdef{#1}{\textsc{\lowercase{#1}}\xspace}}
\tsc{WGM}
\tsc{QE}
\tsc{EP}
\tsc{PMS}
\tsc{BEC}
\tsc{DE}

\newcommand{\given}[1][]{\:#1\vert\:}
\newcommand{\epc}{Euler--Poincar\'e}
\newcommand{\erf}{\text{erf}}

\begin{document}
\let\WriteBookmarks\relax
\def\floatpagepagefraction{1}
\def\textpagefraction{.001}
\shorttitle{Local Morphometry of Closed, Implicit Surfaces}
\shortauthors{Besler et~al.}

\title [mode = title]{Local Morphometry of Closed, Implicit Surfaces}

\author[1,2]{Bryce A Besler}
\author[1,2]{Tannis D. Kemp}
\author[1,2]{Andrew S. Michalski}
\author[2,3]{Nils D. Forkert}
\author[1,2]{Steven K. Boyd}

\address[1]{McCaig Institute for Bone and Joint Health, University of Calgary, Calgary, Canada}
\address[2]{Department of Radiology, University of Calgary, Calgary, Canada}
\address[3]{Hotchkiss Brain Institute, University of Calgary, Calgary, Canada}

\begin{abstract}
Anatomical structures such as the hippocampus, liver, and bones can be analyzed as orientable, closed surfaces.
This permits the computation of volume, surface area, mean curvature, Gaussian curvature, and the \epc~characteristic as well as comparison of these morphometrics between structures of different topology.
The structures are commonly represented implicitly in curve evolution problems as the zero level set of an embedding.
Practically, binary images of anatomical structures are embedded using a signed distance transform.
However, quantization prevents the accurate computation of curvatures, leading to considerable errors in morphometry.
This paper presents a fast, simple embedding procedure for accurate local morphometry as the zero crossing of the Gaussian blurred binary image.
The proposed method was validated based on the femur and fourth lumbar vertebrae of 50 clinical computed tomography datasets.
The results show that the signed distance transform leads to large quantization errors in the computed local curvature.
Global validation of morphometry using regression and Bland-Altman analysis revealed that the coefficient of determination for the average mean curvature is improved from 93.8\% with the signed distance transform to 100\% with the proposed method.
For the surface area, the proportional bias is improved from -5.0\% for the signed distance transform to +0.6\% for the proposed method.
The \epc~characteristic is improved from unusable in the signed distance transform to 98\% accuracy for the proposed method.
The proposed method enables an improved local and global evaluation of curvature for purposes of morphometry on closed, implicit surfaces.
\end{abstract}




\maketitle

\section{Introduction}

Most anatomical structures are orientable, closed surfaces.
Examples include the hippocampus, trabecular bone, and liver.
Since these surfaces are orientable and closed, they permit an embedding, enclosing a defined volume.
This paper is concerned with measuring morphological properties of such surfaces.
While this work was developed in the context of bone-related research, it is applicable to any object representable as a closed surface.

Methods for measuring the volume, area, and curvatures of surfaces is well described.
When the surface is represented as a triangulated mesh, curvatures can be estimated at each vertex by using geometric information from neighboring vertices \cite{goldfeather2004novel,rusinkiewicz2004estimating,flynn1989reliable}, volume by summing the signed volume of each face~\cite{zhang2001efficient}, and area by summing the area of each triangle~\cite{zhang2001efficient,alyassin1994evaluation}.
Alternatively, the surface can be represented implicitly as the level set of an embedding~\cite{osher1988fronts}.
Doing so permits estimating the curvatures locally based on the embedding gradients, while volume, area, and total curvatures can be computed from volume integrals of the embedding~\cite{sethian1999level,chan2001active}.

There are advantages to using the implicit representation over the parametric mesh representation for morphometric analyses.
The first advantage is that spatial gradients of the embedding are well defined, avoiding the need to smooth or fit the surface \cite{goldfeather2004novel,rusinkiewicz2004estimating,flynn1989reliable}.
Second, morphometry can be measured during curve evolution problems where topology can change without explicit splitting and merging techniques~\cite{osher1988fronts}.
This has been the primary feature that made level set methods popular, used extensively in computational fluid dynamics~\cite{peng1999pde,sussman1994level}, object segmentation~\cite{chan2001active,caselles1993geometric,vese2002multiphase}, and biophysical simulations~\cite{besler2018bone}.
The one disadvantage is that implicit representations can require large amounts of memory to store and process.

However, there is an artifact that occurs during embedding that prevents the application of these methods to study anatomical structures.
More precisely, anatomical structures are typically represented as binary images, which are embedded using the signed distance transform~\cite{danielsson1980euclidean}.
However, due to a quantization error in the distance transform of binary images, gradients in the image are very noisy\sloppy~\cite{besler2020artifacts}.
Thus, measures of local mean and Gaussian curvature are poorly estimated based on these embeddings.
This work is principally concerned with demonstrating the unsuitability of the signed distance transform and providing an alternative.
It summarizes morphometrics for orientable, closed surfaces and provides an embedding method suitable for their computation.
The method is local, meaning that the morphometrics can be evaluated at arbitrary locations along the surface.

\begin{figure}
  \centering
  \includegraphics[width=\linewidth]{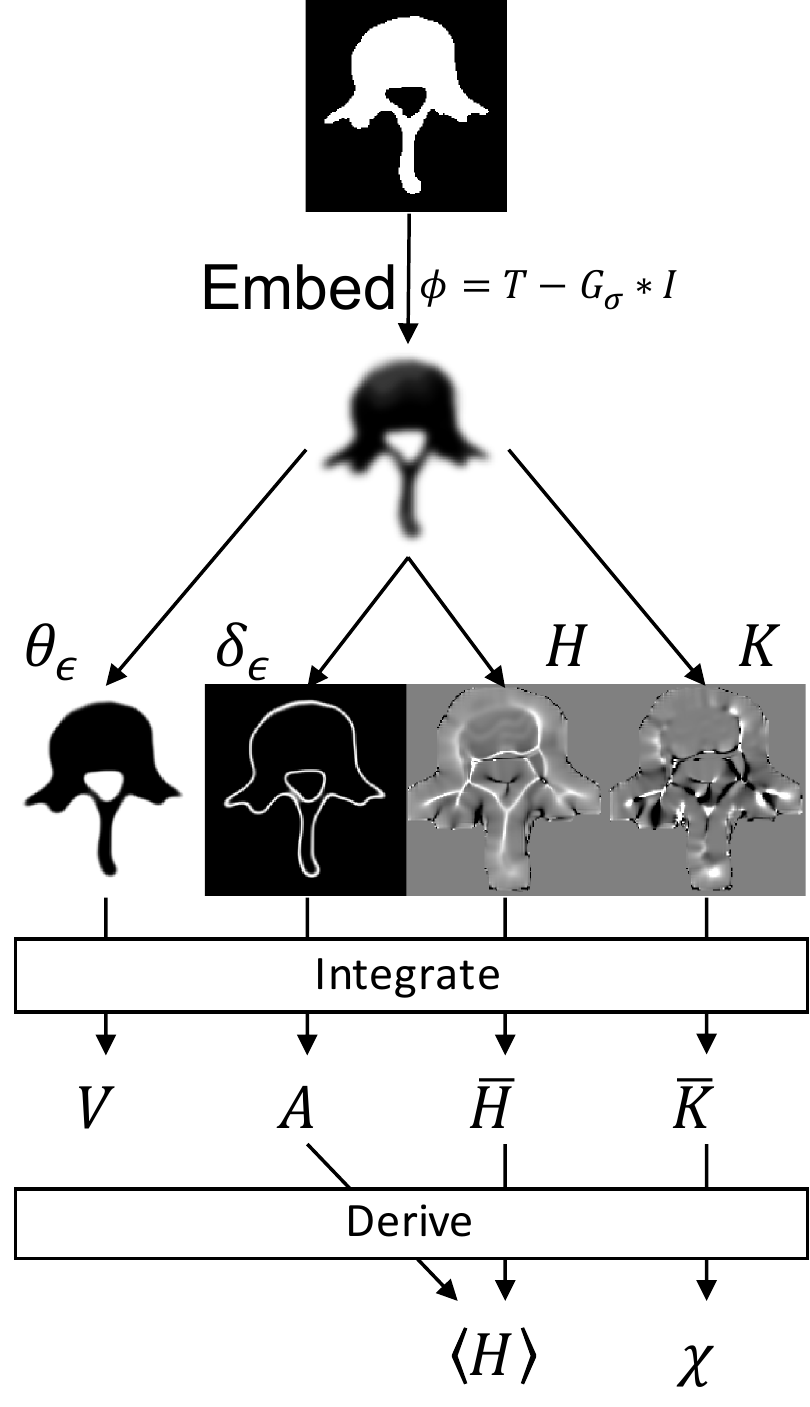}%
  \caption{Workflow for local morphometry of closed surfaces demonstrated on a lumbar vertebra. Clipping in the curvature images originates from the finite support of the Gaussian filter.}
  \label{fig:overview}
\end{figure}

\begin{table}
  \centering
  \begin{tabular}{ccc} 
    \hline
    Parameter & Units & Selection \\
    \hline
    $\epsilon$ & $[-]$ & Equation~\ref{eqn:epsilon} \\
    $\sigma$ & [$\si{\milli\metre}$] & $<$ Structure Thickness \\
    $t$ & [$\si{\milli\metre}$] & $> \Delta x$ \\
    $T$ & $[-]$ & $0.5$ \\
    \hline
  \end{tabular}
  \caption{Summary of the method parameters. Only two parameters are needed: Gaussian blur standard deviation, $\sigma$, and the thickness over which to compute area elements, $t$. $\Delta x$ denotes the voxel resolution.}
  \label{tab:parameters}
\end{table}

\section{Morphometry of Closed Surfaces}
\begin{figure}
  \centering
  \includegraphics[width=\linewidth]{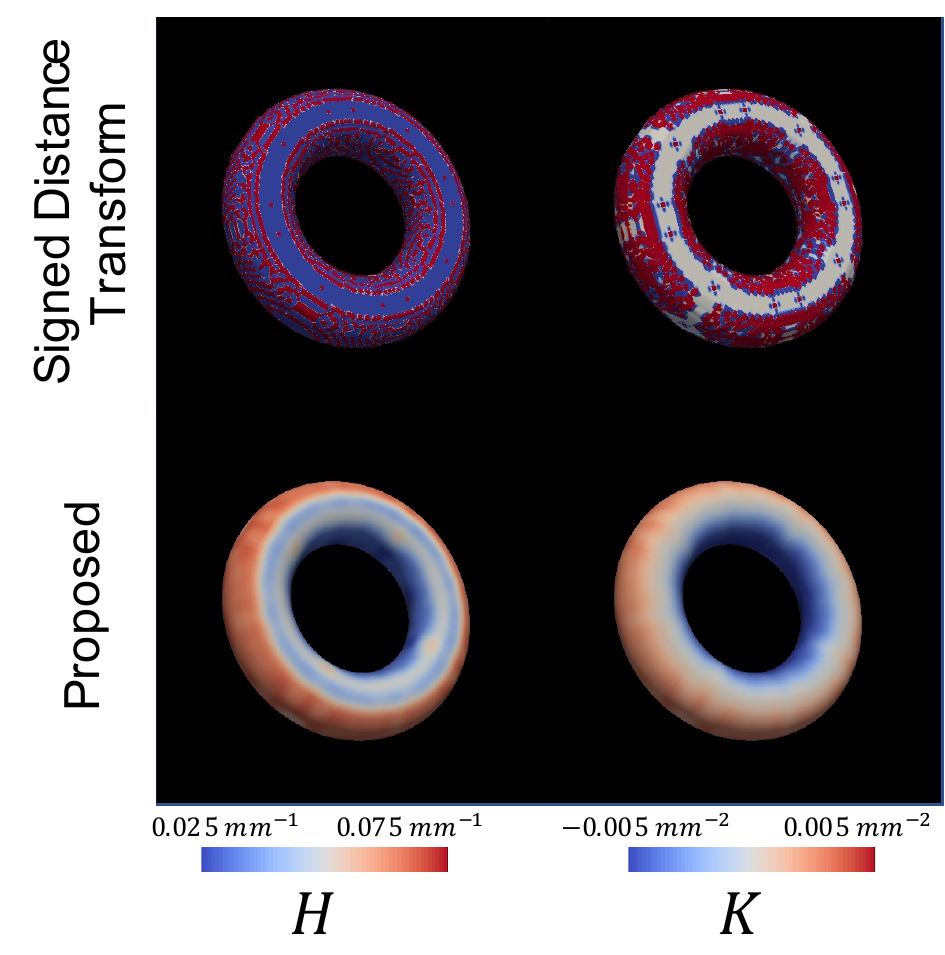}%
  \caption{Local differences between the signed distance transform and proposed method applied to a binary torus (inner radius = $\SI{20}{\milli\metre}$ outer radius = $\SI{40}{\milli\metre}$).}
  \label{fig:torus}
\end{figure}

An overview of the method is given in Figure~\ref{fig:overview} and a summary of the method parameters is given in Table~\ref{tab:parameters}.
The method relies on the local evaluation of the mean and Gaussian curvature as well as volume integrals to derive global morphometrics.
Since the computation of curvatures is local, they can be visualized across the surface.
A motivating example for this work is given in Figure~\ref{fig:torus} where it is demonstrated that a signed distance transform produces enormous errors in local curvature, whereas the proposed method produces smoother and more realistic results.

\subsection{Mathematical Preliminaries}
\subsubsection{Differential Geometry}
Define an orientable, closed, two-dimensional surface $C :  \mathbb{R}^2 \rightarrow \mathbb{R}^3$.
Being closed and orientable allows the surface to define a volume.
Two principal curvatures, $\kappa_1$ and $\kappa_2$, exist at each point on the surface measuring the least and greatest curvature at that point.
The Gaussian ($K$) and mean ($H$) curvatures are defined as the product and average of the principal curvatures.
\begin{eqnarray}
  K &=& \kappa_1 \kappa_2 \\
  H &=& \frac{1}{2} (\kappa_1 + \kappa_2)
\end{eqnarray}
Mean curvature is an extrinsic property of the surface, which can be understood intuitively as the divergence of the normals.
The Gaussian curvature is an intrinsic property of the surface, which can be understood intuitively as the amount of shrinking or expanding that occurs walking along the surface.
It is related to the topology of the surface as is elucidated below.

\subsubsection{Topology}
Considering stretching, bending, and compressing the surface, it is possible to mold a sphere into a femur without cutting or gluing the object.
The study of surfaces related by an isomorphism is called topology.
An important measure in topology is the \epc~characteristic.
If an object is viewed as a graph or mesh, the \epc~characteristic $\chi$ can be computed from the vertices $V$, edges $E$, and faces $F$ of the mesh:
\begin{equation}
  \chi = V - E + F
\end{equation}
When viewed continuously, the \epc~characteristic is related to the genus $g$(colloquially, the number of holes) of a surface.
\begin{equation}
  \chi = 2 - 2 g
\end{equation}
Both, the \epc~characteristic and genus, are topological invariants, meaning they do not change with the bending and stretching of the surface, only with cutting or gluing.
By example, to mold a femur into a vertebra, a hole must be created corresponding to the foramen (colloquially, spinal cord hole).

\subsubsection{Gauss-Bonnet Theorem}
Remarkably, local measures of curvature can be related to their topology.
More precisely, the Gauss-Bonnet theorem states that the Gaussian curvature summed across a surface is equal to the \epc~characteristic.
\begin{equation}
  \int_M K dA = 2 \pi \chi
\end{equation}
In this work, the Gauss-Bonnet theorem will be used to measure the \epc~topological invariant from local Gaussian curvature.

\subsection{Embedding of Closed Surfaces}
The problem of embedding a closed surface is described. Consider a binary image $I: \Omega \rightarrow \{0,1\}$ to be embedded, where $\Omega \subset \mathbb{Z}^n$ is the discrete domain.
An embedding $\phi$ is sought such that it recovers the underlying binary image:
\begin{equation}
  \label{eqn:recovery}
  \theta(-\phi) = I
\end{equation}
where $\theta$ is the Heaviside function.
The nomenclature common in statistics, $\theta$, is used for the Heaviside to avoid confusion with mean curvature.
Furthermore, the surface is recoverable as the zero level set of the embedding:
\begin{equation}
  \label{eqn:level_set}
  C(x) = \{ x \given \phi(x) = 0\}
\end{equation}
The surface can be any level set of the embedding but will be taken as the zero level set in this work.
The problem is under-constrained and does not permit a unique embedding.
As a convention, this work considers the inside of the surface as having a negative embedding.
An embedding is always possible for a closed and orientable surface.

The embedding $\phi$ is a non-parametric representation containing the same information as $C$.
However, it can be much easier to work with $\phi$ computationally than $C$ because of issues of parametrization.
The success of the level set method~\cite{osher1988fronts} is largely due to the ease of working with the embedding while being able to recover the surface at a later time.

\subsubsection{Signed Distance Transform}
The most commonly used embedding is the signed Euclidean distance transform~\cite{rosenfeld1966sequential,danielsson1980euclidean}.
This transform assigns a value to every point $x$ in the image based on its signed distance $d$ from the surface $C$:
\begin{equation}
  \label{eqn:sdt}
  \phi(x) = \pm d(x, C)
\end{equation}
The embedding is unique given the additional constraint that the magnitude gradient of the embedding equals $+1$.
The signed distance transform is a computationally fast method of embedding a binary image~\cite{danielsson1980euclidean}.
However, the distance transform of sampled signals produces a quantized representation of the true signal~\cite{besler2020artifacts}.
As a result, gradients are extremely noisy and independent of image spacing.
Furthermore, reinitialization methods~\cite{peng1999pde,sussman1994level} to overcome this problem converge slowly~\cite{besler2020artifacts} making them impractical for removing quantization errors.

\subsubsection{Proposed Embedding Technique}
A different embedding is proposed in this work based on a Gaussian blur of the binary image.
The image intensities are shifted by a threshold $T \in [0, 1]$ such that the zero crossing corresponds to the binary surface:
\begin{equation}
  \label{eqn:embedding}
  \phi = T - G_\sigma * I
\end{equation}
where $G_\sigma$ denotes a Gaussian filter of standard deviation $\sigma$ and $*$ denotes the convolution operator.
There are two parameters to this embed
ding, the threshold and standard deviation.
The threshold should be selected as $0.5$ to preserve the localization of flat surfaces and the standard deviation should be selected larger than the size of a voxel but not larger than the structure.
The optimal amount of smoothing is application specific.
Gaussian blurring a binary image to generate a surface mesh using Marching Cubes is a common task in image processing~\cite{lorensen1987marching}.

\begin{figure}
  \centering
  \includegraphics[width=\linewidth]{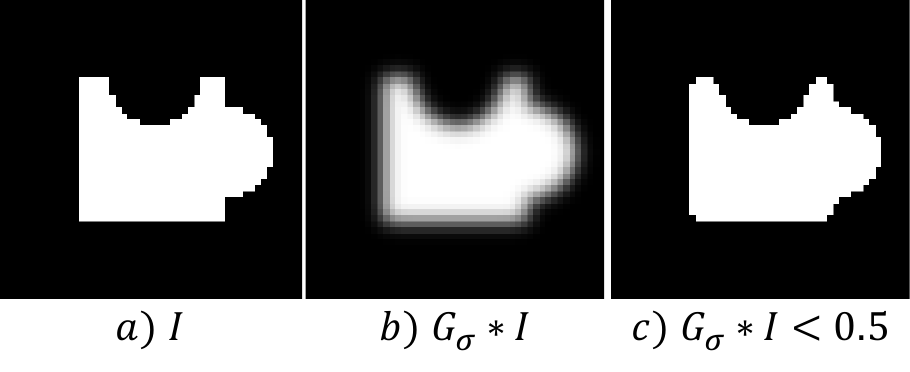}%
  \caption{Gaussian blurring modifies the underlying object like mean curvature flow.}
  \label{fig:gauss}
\end{figure}

Properties of the proposed embedding technique should be made explicit.
First, the proposed method modifies the binary image.
That is, the Heaviside of the embedding does not recover the original binary image exactly.
Areas of concavity shrink and areas of convexity expand (Figure~\ref{fig:gauss}).
Second, the embedding technique does not produce a signed distance image.
If a signed distance signal is needed, reinitialization~\cite{peng1999pde,sussman1994level,kimmel1996sub} can be performed on the embedding.
Finally, the resulting image has intensities in the range $[-0.5,0.5]$.
Within this context, it should be noted that many binary images are stored as the largest value in their dynamic range ($127$ for a signed char, $255$ for an unsigned char) and should be flattened to $\{0,1\}$ before embedding as described above.

\subsubsection{Relation to Mean Curvature Smoothing}
While Gaussian blurring smooths the binary image, we would like to know how that translates into surface smoothing.
Prior literature is used to show that Equation~\ref{eqn:embedding} leads to mean curvature smoothing of the surface.

First, consider the heat flow of the image:
\begin{numcases}{\hspace{-20pt}} 
  I_\tau = \Delta I & $\text{on }\Omega \times (0, \infty)$ \\
  I(x,0) = I_0 & $\text{on }\Omega \times 0$
\end{numcases}
where $\tau$ is a time-like parameter, $\Delta = \nabla^2$ is the scalar Laplacian, and $I_0$ is the original binary image.
It is well known that the solution of the heat equation is Gaussian convolution~\cite{witkin1984scale,koenderink1984structure}:
\begin{equation}
  I(x,\tau) = G_\tau(x) * I_0(x)
\end{equation}
with $2\tau = \sigma^2$.
The image $I(x,\tau)$ is related to Equation~\ref{eqn:embedding} by shifting the embedding such that the $T$ level set in $I(x,\tau)$ is the zero level set in $\phi$.

Importantly, the heat flow and level set shift corresponds to the BMO (Bence-Merriman-Osher) algorithm in computational physics for simulating mean curvature flow~\cite{merriman1992diffusion}.
BMO simulates mean curvature flow by blurring a binary image using the heat equation and rebinarizing the field with a threshold at $0.5$.
Evans (Theorem 5.1, \cite{evans1993convergence}) proved that if $u$ is the viscous solution from mean curvature flow and $I(x,\tau)$ the solution from the diffusion equation, the two methods are equivalent in the limit of small $\tau$.

As Equation~\ref{eqn:embedding} has the effect on the surface of mean curvature flow, general principles can be known about how the technique modifies the surface~\cite{evans1991motion}.
First, mean curvature flow is the gradient descent of the first variation of surface area.
As such, the surface area of the object will be greatly affected by blurring.
Furthermore, areas of high curvature, such as at the tip of the transverse process in lumbar vertebrae, experience more change than flat areas, such as the articulating surface of the vertebral body.
Second, minimal surfaces where mean curvature is zero everywhere will experience no local shift in the surface location.
Finally, the topology of the surface can change corresponding to singularities in mean curvature flow.

\subsection{Morphometry of Embedded Surfaces}
Attention is now placed on measuring geometric and topological parameters from an embedding.

\subsubsection{Measures of Volume and Area}
Methods for measuring the volume and area of an implicit surface have been known for some time~\cite{sethian1999level,chan2001active} and are summarized here.
Briefly, the volume of the surface can be determined by summing up all volume elements inside the surface.
This can be well-defined as the integral of the Heaviside function of the embedding:
\begin{equation}
  \text{Volume}(\phi) = \int_\Omega \theta(-\phi) dV
\end{equation}
where $dV$ is the volume of a volume element.
By considering area as the variation of volume, the area can be defined equally as well:
\begin{equation}
  \label{eqn:area}
  \text{Area}(\phi) = \int_\Omega \delta(\phi) \lvert \nabla \phi \rvert dV
\end{equation}
where $\delta = \lvert \nabla \theta \rvert$ is the Dirac delta function and $\nabla$ is the differential operator.

\subsubsection{Measuring Local Curvature}
For an embedding, the Gaussian and mean curvature are typically computed first and then principal curvatures derived~\cite{sethian1999level}.
Mean curvature is computed as one half the divergence of the surface normal:
\begin{eqnarray}
  N &=& \frac{\nabla \phi}{\lvert \nabla \phi \rvert} \\
  H &=& \frac{1}{2} \nabla \cdot \left( \frac{\nabla \phi}{\lvert \nabla \phi \rvert} \right)
\end{eqnarray}
where $N$ denotes the unit normal vector, $\nabla$ is the gradient operator, $\lvert\cdot\rvert$ is the $\ell^2$ norm, and $H$ is mean curvature.
The one-half factor is not typically used in the literature on level set methods and mean curvature flow~\cite{osher1988fronts,sethian1999level}.
It comes from averaging the principal curvatures of the surface, of which there are two on two-dimensional surfaces.
This brings the computation of mean curvature of the embedding equal to the mean curvature as defined in differential geometry.
In higher dimensions, the factor would be one divided by one less the dimension of the embedding domain.

Similarly, the Gaussian curvature can be defined from the level set embedding in terms of first and second derivatives of the embedding:
\begin{equation}
  K = -\frac{\begin{vmatrix}
    \phi_{xx} & \phi_{xy} & \phi_{xz} & \phi_x \\
    \phi_{yx} & \phi_{yy} & \phi_{yz} & \phi_y \\
    \phi_{zx} & \phi_{zy} & \phi_{zz} & \phi_z \\
    \phi_x & \phi_y & \phi_z & 0
    \end{vmatrix}}{\lvert \nabla \phi \rvert^4}
\end{equation}
where $\lvert \cdot \rvert$ denote the matrix determinant and $K$ is Gaussian curvature.

While not used in this work, the principal curvatures can be computed from the mean and Gaussian curvature.
\begin{equation}
  \kappa_1, \kappa_2 = H \pm \sqrt{H^2 - K}
\end{equation}

\subsubsection{Computing Total Curvature}
Based on the definition of area, a way of computing integrals along the surface is defined.
Consider some quantity $Q$ to be integrated over a surface.
This integral can be generalized using the definition of area given in Equation~\ref{eqn:area}:
\begin{equation}
  \label{eqn:surface_integrals}
  \int_M Q dA = \int_\Omega Q \delta(\phi) \lvert \nabla \phi \rvert dA
\end{equation}
The form of this integral allows surface integrals to be performed in general. Thus, the total mean ($\bar{H}$) and Gaussian curvature ($\bar{K}$) can be computed:
\begin{eqnarray}
  \bar{H} &=& \int_\Omega H \delta(\phi) \lvert \nabla \phi \rvert dA \\
  \bar{K} &=& \int_\Omega K \delta(\phi) \lvert \nabla \phi \rvert dA
\end{eqnarray}
Now that area, volume, total mean curvature, and total Gaussian curvature are defined, other morphometric quantities can be derived.
First, the average mean curvature $\langle H \rangle$ can easily be defined by dividing by the total area.
\begin{equation}
  \langle H \rangle = \frac{\bar{H}}{A}
\end{equation}
By the Gauss-Bonnet theorem, the \epc~characteristic can also be computed.
\begin{equation}
  \chi = \frac{\bar{K}}{2\pi}
\end{equation}
Volume, area, average mean curvature, and the \epc~characteristic are the most natural global descriptors of surfaces.

\subsection{Implementation Considerations}
\subsubsection{Numerical Approximations}
A numerical approximation is needed for the Heaviside and Dirac delta functions.
Many are available~\cite{chan2001active} but the finite support sine approximation is used in this work:
\begin{eqnarray}
  \theta_\epsilon(x) &=& \begin{cases}
    \frac{1}{2} \left[1 + \frac{x}{\epsilon} + \frac{1}{\pi} \sin\left(\frac{\pi x}{\epsilon}\right) \right] & \lvert x \rvert \leq \epsilon \\
    1 & x > \epsilon \\
    0 & x < -\epsilon
  \end{cases} \\
  \delta_\epsilon(x) &=& \begin{cases}
    \frac{1}{2\epsilon} \left[ 1 + \cos\left(\frac{\pi x}{\epsilon}\right)\right] & \lvert x \rvert \leq \epsilon \\
    0 & \lvert x \rvert > \epsilon
  \end{cases}
\end{eqnarray}
The finite support is advantageous as it is conceptually easy to design with in comparison to infinite support approximations such as the hyperbolic tangent.
Outside $2\epsilon$, the response is zero or one.

The regularization parameter $\epsilon$ should be selected larger than one voxel edge and less than the support of the Gaussian filter.
For the distance transform, $\epsilon$ is often selected as some multiple of image resolution.
However, the proposed embedding technique does not permit such a simple method of parameter selection since the embedding no longer encodes physical space.
Instead, if the intent is to average over some physical thickness $t$, $\epsilon$ should be selected as:
\begin{equation}
  \label{eqn:epsilon}
  \epsilon = \frac{1}{2} \erf\left(\frac{t}{2\sqrt{2}\sigma}\right)
\end{equation}
where $\sigma$ is the filter standard deviation and $\erf\left(\cdot\right)$ is the error function.
Derivation of Equation~\ref{eqn:epsilon} is given in Appendix~\ref{app:regulariz_selection}.
The parameter $t$ is termed the regularization thickness to denote it has dimensions of length and can be interpreted as a physical size.

Furthermore, the integrals of the form of Equation~\ref{eqn:surface_integrals} need to be numerically approximated.
Simply summing across the image and multiplying by the product of spacing works well.
Alternatively, Simpson’s rule can be applied along each direction in the image.
This leads to a slight improvement of the volume measurement for the monotonic Heaviside function where single sided approximation of the Riemann integral leads to a systematic error.
This work uses Simpson’s rule for completeness.

\subsubsection{Finite Differences}
All infinitesimal differences are approximated with finite differences.
This work uses fourth-order accurate central differences for the first, second, and mixed derivatives.
At least second-order accurate differences are required~\cite{coquerelle2016fourth} and all derivatives should be of the same order.
No boundary conditions are defined for the problem, and they are not needed.
Either the image can be padded based on features of the embedding (details in Section~\ref{subsubsec:image_boundary}) or the finite difference stencil can be shifted near the edges of the image.
Finite difference equations are given in Appendix~\ref{app:stencils}.

\subsubsection{Image Boundary}
\label{subsubsec:image_boundary}
Care should be taken at the boundary of the image for two reasons.
First, the surface may be clipped by the edge of the image, modifying morphometry.
Second, the formula for the \epc~characteristic includes an additional term when the surface is clipped, which is not obvious to compute.
The image can be padded to avoid boundary issues.
If the given data is a binary image, background voxels can be padded before embedding.
If the given data is an embedding, padding can be selected as appropriate for the embedding method.
For distance transforms where the inside of the curve is negative, this would be padding with a positive number.
For the proposed method, this would be padding with a constant of 0.5.

Still, some structures are artificially clipped by the imaging protocol.
For instance, abdominal clinical computed tomography clips the femur at the lesser trochanter and extremity imaging clips long bones to the scanner field of view.
Closing these surfaces by padding background voxels is required.
Once closed, clipping location will still affect the measured outcomes and care should be taken to standardize clipping in a study.

\subsubsection{Visualization and Histograms}
One advantage of the proposed method is that we can visualize curvatures on the surface.
If a volume renderer~\cite{drebin1988volume} is used, opacity can be mapped by a regularized Heaviside function and color mapped by a transfer function of the embedding.
Alternatively, the object can be meshed using marching cubes at the zero iso-contour~\cite{lorensen1987marching} and finite difference stencils placed on the mesh vertices to compute mean and Gaussian curvature.
As gradients are well defined, both methods give excellent visualizations.
Marching cubes is used in this work as it permits the computation of histograms from vertices.

\begin{figure}
  \centering
  \includegraphics[width=\linewidth]{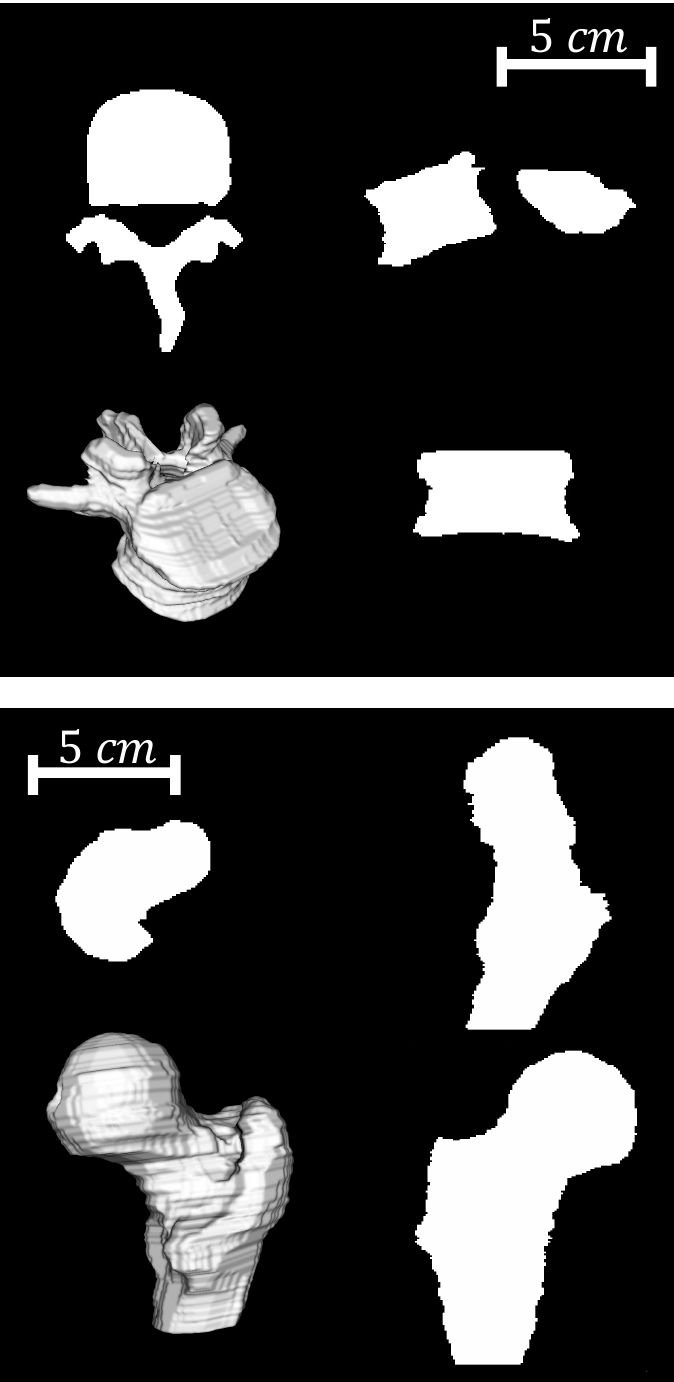}%
  \caption{Visualization of the dataset. L4 vertebra are fully imaged while the proximal femur is clipped at an operator defined level roughly corresponding to the  lesser trochanter. Manual segmentation artifacts are evident. Starting top left and going clockwise, the images are an axial slice, sagittal slice, coronal slice, and 3D render. Scale bar is for the 2D visuals.}
  \label{fig:data}
\end{figure}

\section{Experiments}
Experiments are conducted to demonstrate the unsuitability of the signed distance transform, gain intuition on the parameters of the proposed embedding and morphometry technique, and validate the morphometry against existing method.

50 abdominal clinical computed tomography images are used.
The scan volume started at the T12 vertebrae and ended at the lesser trochanter.
The right femur and fourth lumbar vertebra were manually segmented from each dataset.
33 (66\%) of subjects were male. Reported median [min -- max], age was 61.5 [50.0 – 102.0] years, in-plane resolution was 0.703 [0.580 -- 0.977] $\si{\milli\meter}$, and slice thickness was 0.625 [0.624 -- 1.000] $\si{\milli\meter}$.
A median (by volume) segmented femur and vertebra are displayed in Figure~\ref{fig:data}.
Small artifacts due to manual segmentation along the axial direction are evident. Additional details on the data can be found in a previous study~\cite{michalski2021opportunistic}.

\subsection{Necessity of the Embedding Technique}
One vertebra was used to visually demonstrate that the signed distance transform is insufficient for local morphometry.
This subject had an in-plane resolution of 0.639 mm and a slice thickness of $\SI{0.625}{\milli\meter}$.
The binary image was embedded with the signed distance transform~\cite{danielsson1980euclidean} and with the proposed method ($\sigma = \SI{1.0}{\milli\meter}$).
The surface was extracted using the marching cubes method.
Mean and Gaussian curvatures were estimated at each vertex and visualized across the mesh triangles.
The histogram for mean and Gaussian curvature were generated from the vertices.

\subsection{Structural Changes of the Proposed Embedding Technique}
\label{subsec:structural}
The next experiment aimed at investigating the change in structure as a consequence of blurring.
Therefore, the right femur and L4 vertebrae of one subject were used.
This subject had an in-plane resolution of $\SI{0.672}{\milli\metre}$ and a slice thickness of $\SI{1.000}{\milli\metre}$.
Embedding was repeated for 100 Gaussian standard deviations spaced uniformly from $\SI{1.0}{\milli\meter}$ to $\SI{5.0}{\milli\metre}$.
The regularization thickness ($t = \SI{1.5}{\milli\meter}$) did not change.
Measured outcomes were volume (V, [$\si{\milli\metre\cubed}$]), area (A, [$\si{\milli\metre\squared}$]), average mean surface curvature ($\langle H \rangle$, [$\si{\per\milli\metre}$]), and \epc~characteristic ($\chi$, [–]).
Embeddings at select standard deviations are visualized for qualitative assessment.

\subsection{Sensitivity of Morphometrics to Regularization Thickness}
Next, the stability of the morphometric calculations to regularization thickness was explored.
The same dataset and same morphometric outcomes were used as described in Section~\ref{subsec:structural}.
Outcomes were plotted for 100 regularization thicknesses spaced uniformly from $\SI{0.5}{\milli\metre}$ to $\SI{10.0}{\milli\metre}$.
The blurring ($\sigma = \SI{2.5}{\milli\metre}$) was kept constant for this experiment.

\begin{figure*}
  \centering
  \includegraphics[width=\linewidth]{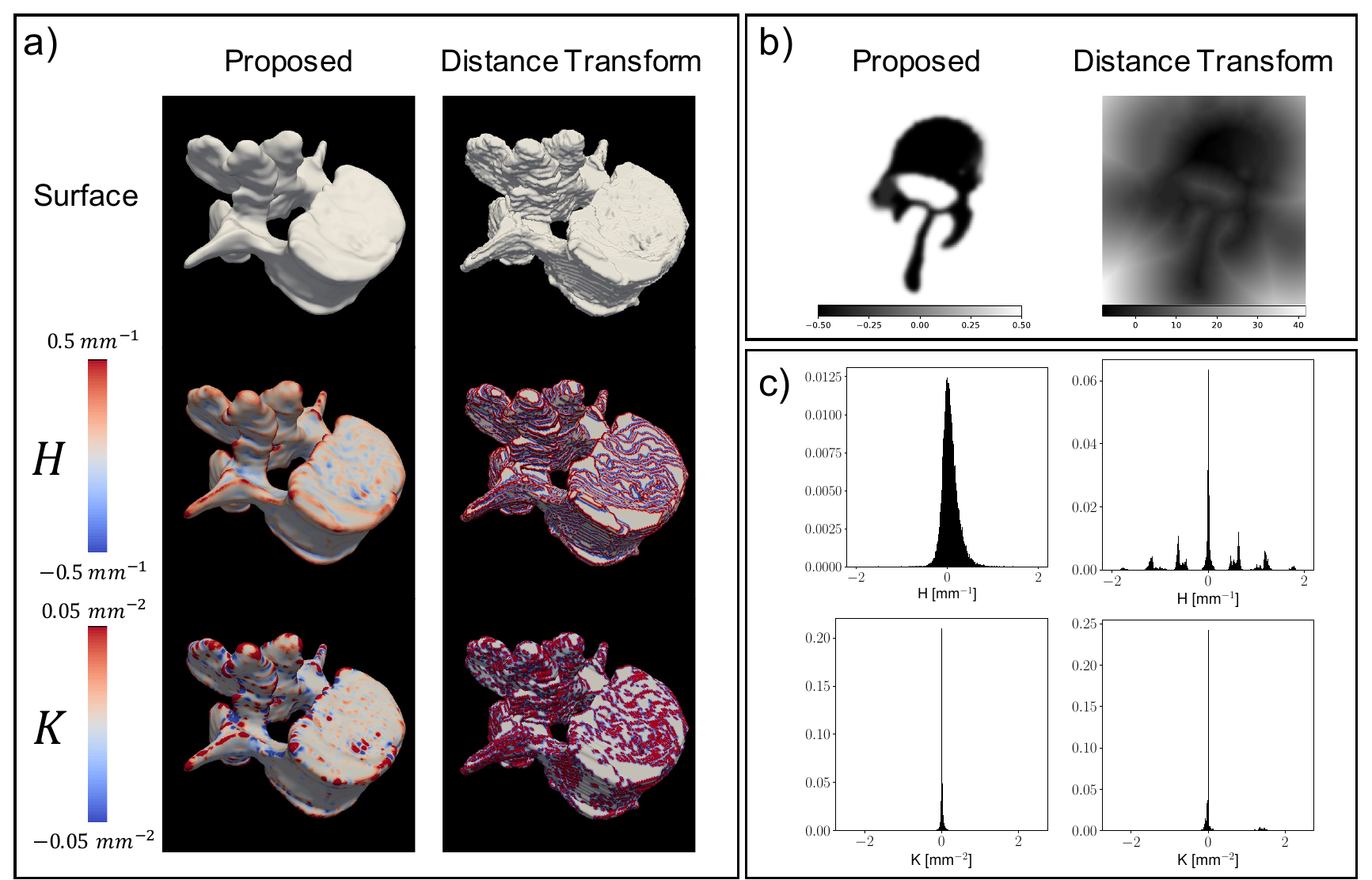}%
  \caption{Distance transforms produce quantization artifacts in the computation of curvature. (\ref{fig:sdt}a) Visualization of the surface extracted from the respective embeddings, (\ref{fig:sdt}b) 2D axial slices through the embeddings, (\ref{fig:sdt}c) Histograms of the curvatures evaluated at the vertices of the extracted surface. The curvature histograms computed from the distance transform demonstrate severe artifacts.}
  \label{fig:sdt}
\end{figure*}

\subsection{Validation of Morphometrics}
The objective of the final experiment was to compare global morphometric outcomes of the proposed method with established methods.
All 50 right femurs and 50 4th lumbar vertebrae were used for comparison. Images were embedded ($\sigma = \SI{2.0}{\milli\metre}$) and morphometry performed with the proposed method ($t = \SI{2.5}{\milli\metre}$).
Measured outcomes were volume (V, [$\si{\milli\metre\cubed}$]), area (A, [$\si{\milli\metre\squared}$]), surface average mean curvature ($\langle H \rangle$, [$\si{\per\milli\metre}$]), and \epc~characteristic ($\chi$, [–]).
The embeddings were re-binarized by thresholding below zero to compare the proposed method with traditional morphometric techniques.
In this way, the underlying structure is the same in both methods.
The binary image was embedded using a signed distance transform and morphometry performed ($\epsilon = \SI{1.25}{\milli\metre}$) to quantify error when using a signed distance transform embedding.

Traditional morphometry was computed on the binary image using Image Processing Language (IPL v5.42, SCANCO Medical AG, Brüttisellen, Switzerland).
\epc~characteristic was computed using an exact method for a 3D binary image~\cite{odgaard1993quantification}, area and volume are estimated by triangulating the surface, and surface average mean curvature is computed using a dilation technique~\cite{hildebrand1997quantification}.

The proposed measures of volume, area, and average mean surface curvature were compared to standard techniques using regression and Bland-Altman analysis~\cite{bland1986statistical}.
The categorical \epc~characteristic was compared using a confusion matrix.
This nuance is outlined in the discussion, but the proposed method gives a continuous outcome for \epc~characteristic while the standard method gives a categorical outcome.
For purposes of comparison, the proposed technique was rounded to an integer.
Analysis was stratified by femur and vertebra. Statistical analysis was performed in R (v4.0.0, The R Foundation for Statistical Computing, Vienna, Austria)~\cite{team2013r}.

\section{Results}

\begin{figure*}
  \centering
  \includegraphics[width=\linewidth]{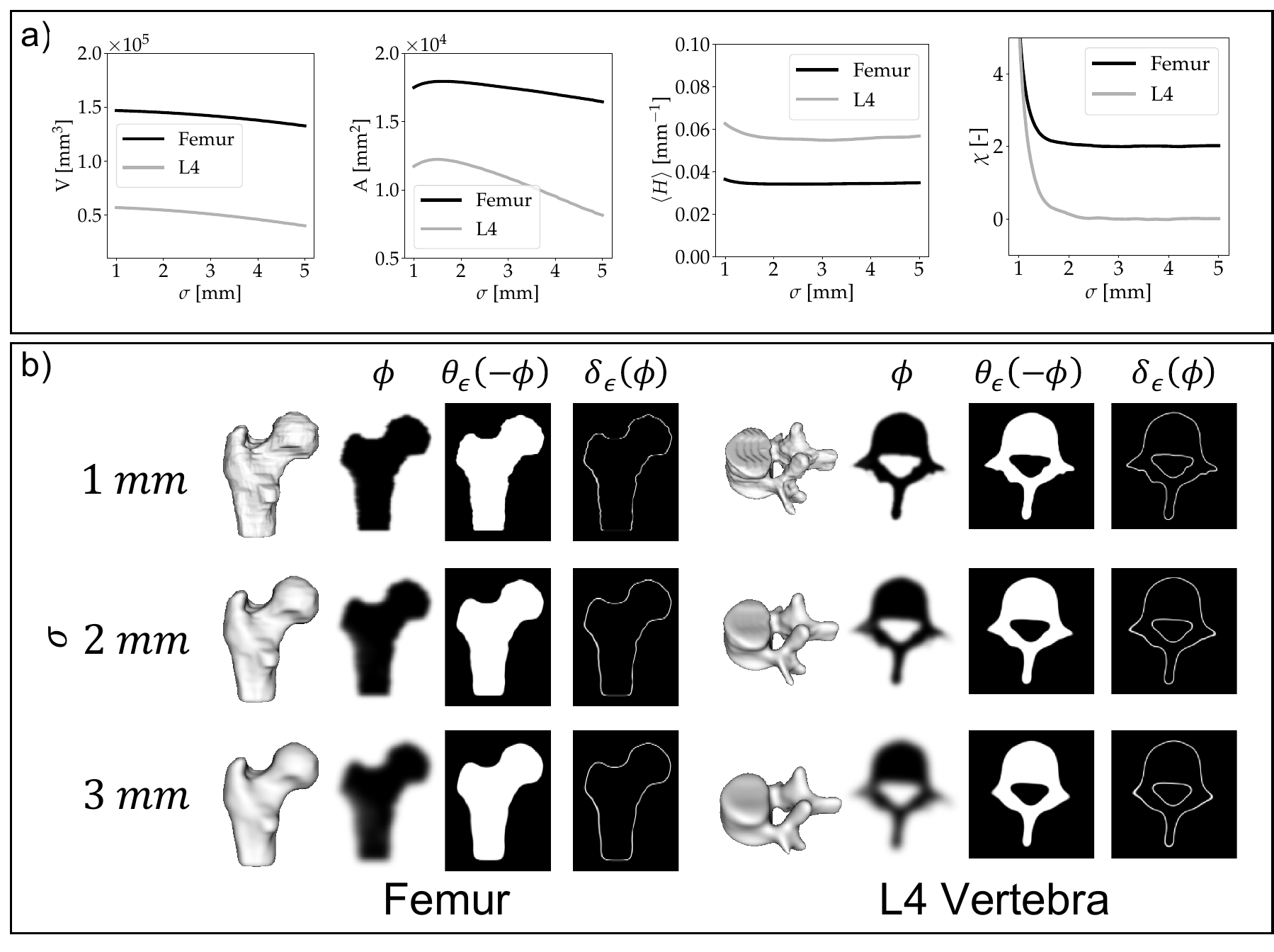}%
  \caption{Structural changes as a function of smoothing. (\ref{fig:smoothing}a) Morphometrics vary as a function of smoothing. (\ref{fig:smoothing}b) Visualizing the surfaces for different smoothing values. $t = \SI{1.5}{\milli\metre}$ throughout.}
  \label{fig:smoothing}
\end{figure*}

\subsection{Necessity of the Embedding Technique}
Surfaces of the proposed and signed distance transform embedding techniques are rendered in Figure~\ref{fig:sdt}a for a selected case.
Overall, the rendered surface is smoother using the proposed method.
In contrast to this, the mean and Gaussian curvature exhibit a large amount of noise when computed from the signed distance transform image.
A 2D slice is taken through the embedding in Figure~\ref{fig:sdt}b. Histograms of the curvatures are displayed in Figure~\ref{fig:sdt}c.
Large curvatures from the signed distance transform are severely quantized compared with the proposed method.
This result is fundamental to the signed distance transform of binary images~\cite{besler2020artifacts}.

\begin{figure*}
  \centering
  \includegraphics[width=\linewidth]{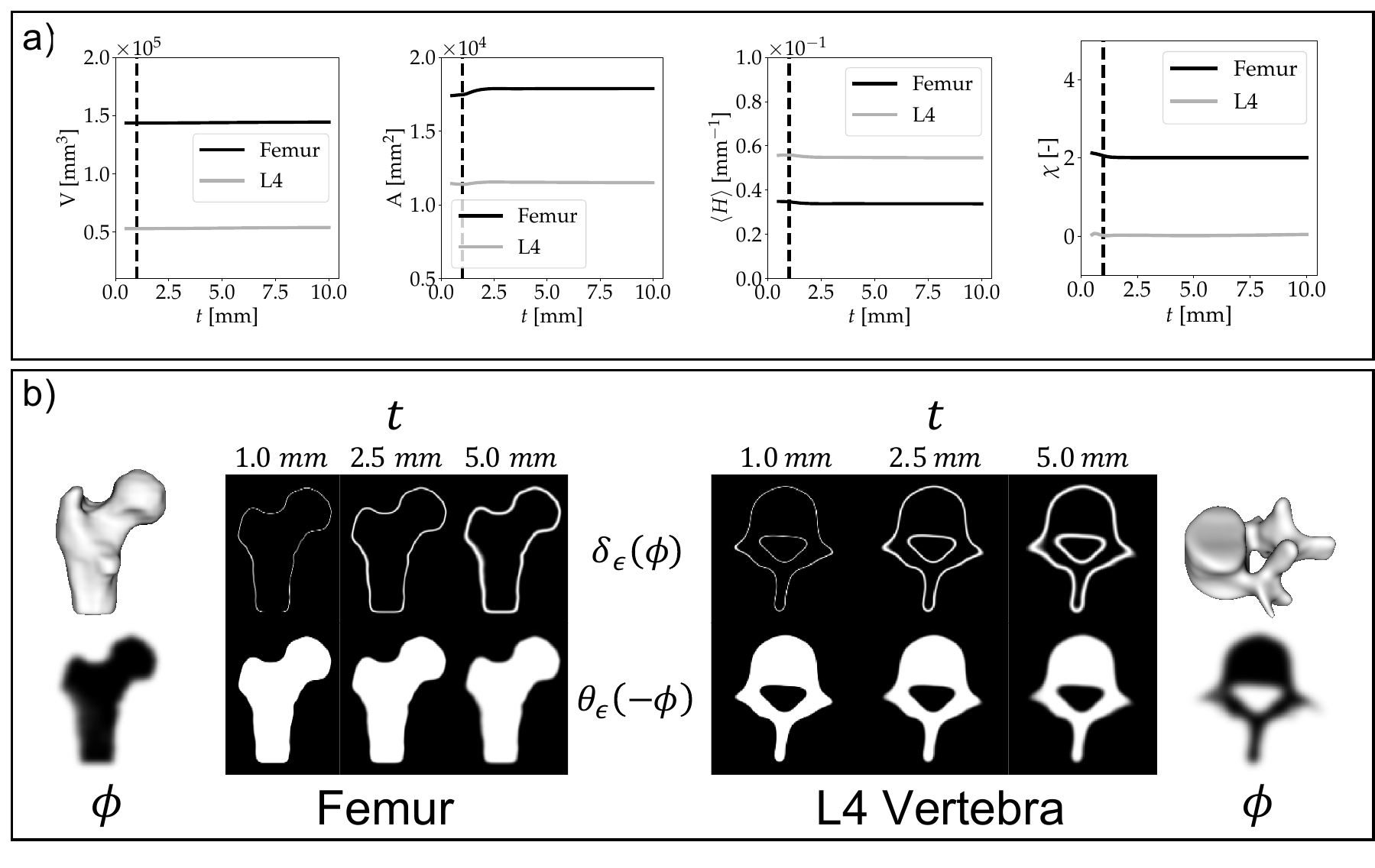}%
  \caption{Sensitivity to regularization thickness. (\ref{fig:regularization}a) Morphometrics as a function of regularization thickness. The dashed vertical line indicates the largest edge length of the dataset’s anisotropic voxel ($\SI{1}{\milli\metre}$). (\ref{fig:regularization}b) Visualizing the Dirac delta and Heaviside response for varying regularization thicknesses. $\sigma = \SI{2.5}{\milli\metre}$ throughout.}
  \label{fig:regularization}
\end{figure*}

\subsection{Structural Changes of the Proposed Embedding Technique}
Changes in morphometry as a function of smoothing are shown in Figure~\ref{fig:smoothing}.
Both, the femur and vertebrae, show unsmooth voxel edges with little smoothing.
At a standard deviation of 3.0 mm, the vertebra is oversmoothed with a bubble-like look.
The \epc~characteristic is poorly estimated for low standard deviations.
Sensitivity of area, volume, and averaged mean surface curvature are consistent with the Gaussian modifying the underlying object.
The transverse processes shrink in Figure~\ref{fig:smoothing}b while small variations along the surface are removed with increasing smoothing.
This is consistent with the decreasing surface averaged mean curvature where areas of high absolute curvature are smoothed more rapidly than areas of low curvature.
Finally, the \epc~characteristic settles near 2 for the femur and 0 for the vertebra.
This is expected as the femur is topologically equivalent to a sphere and the vertebra is topologically equivalent to a torus.
If the smoothing is selected much larger than the thickness of the vertebral arch, a hole may be introduced in the surface, changing the topology.
However, the \epc~characteristic of a vertebrae is not necessarily always zero.
There can be physical damage or anatomical anomalies, which disconnect or form holes in the vertebral arch.
Interestingly, not all vertebrae are topologically equivalent either, with cervical vertebrae having two additional holes corresponding to the transverse foramen.

An important result of this experiment is that small smoothing values still have quantized gradients, evident by the increase in \epc~characteristic.
Such large values of the \epc~characteristic are physically impossible since the only orientable, closed surface with a \epc~characteristic greater than zero is the sphere.
There were no disconnected particles in the image which could have increased the \epc~characteristic past two.
Depending on the object structure relative to image resolution size, the required amount of smoothing may be prohibitive.

When embedding a binary image, it is recommended to select one standard deviation for each tissue class but not necessarily the same standard deviation across tissue classes.
The smoothing should be selected such that morphometry is accurate, but the object is not artificially smoothed for subsequent processing.
Experimental designers using this technique are responsible for quantifying and understanding the structural changes that occur with Gaussian blurring.
In general, smoothing should be selected larger than the image resolution but smaller than the object thickness or holes in the object.

\subsection{Sensitivity of Morphometrics to Regularization}
Changes in morphometry as a function of regularization thickness are plotted in Figure~\ref{fig:regularization}a.
Morphometric outcomes hardly vary as a function of regularization thickness, suggesting that the method is insensitive to regularization thickness.
The \epc~characteristic is the most sensitive outcome exhibiting non-integer values for values of t smaller than a voxel.
Finally, increasing the regularization thickness increases the response size of the Dirac delta and Heaviside responses, as expected (Figure~\ref{fig:regularization}b).

\subsection{Validation of Morphometrics}
\begin{landscape}
  \begin{figure}
    \centering
    \includegraphics[width=0.9\linewidth]{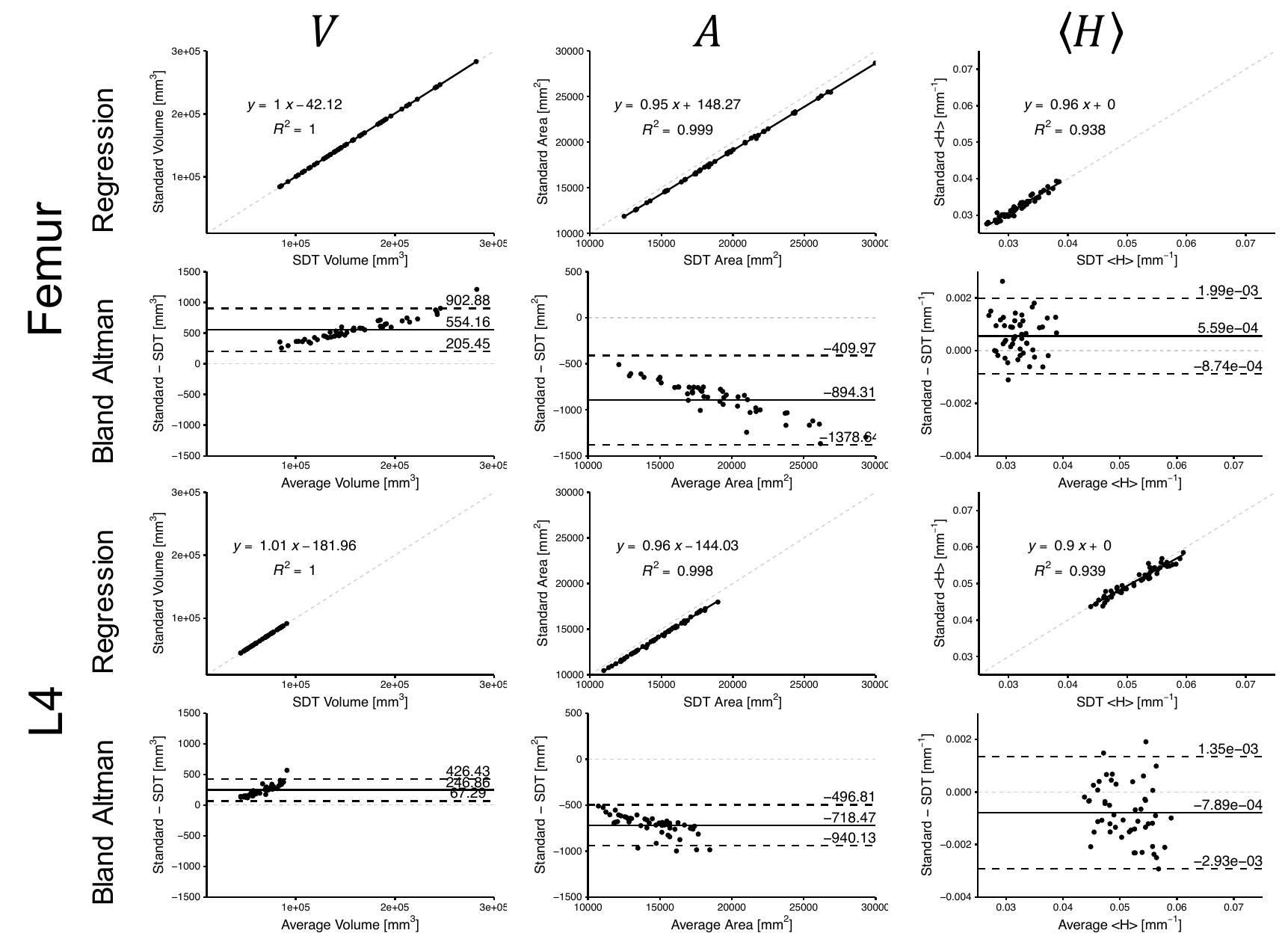}%
    \caption{Regression and Bland-Altman plots demonstrating agreement between the signed distance transform and standard methods for the femur and L4 vertebra (n=50 in each group). Dashed light gray lines denote ideal relationships.}
    \label{fig:regression-ba-sdt}
  \end{figure}
\end{landscape}

\begin{landscape}
  \begin{figure}
    \centering
    \includegraphics[width=0.9\linewidth]{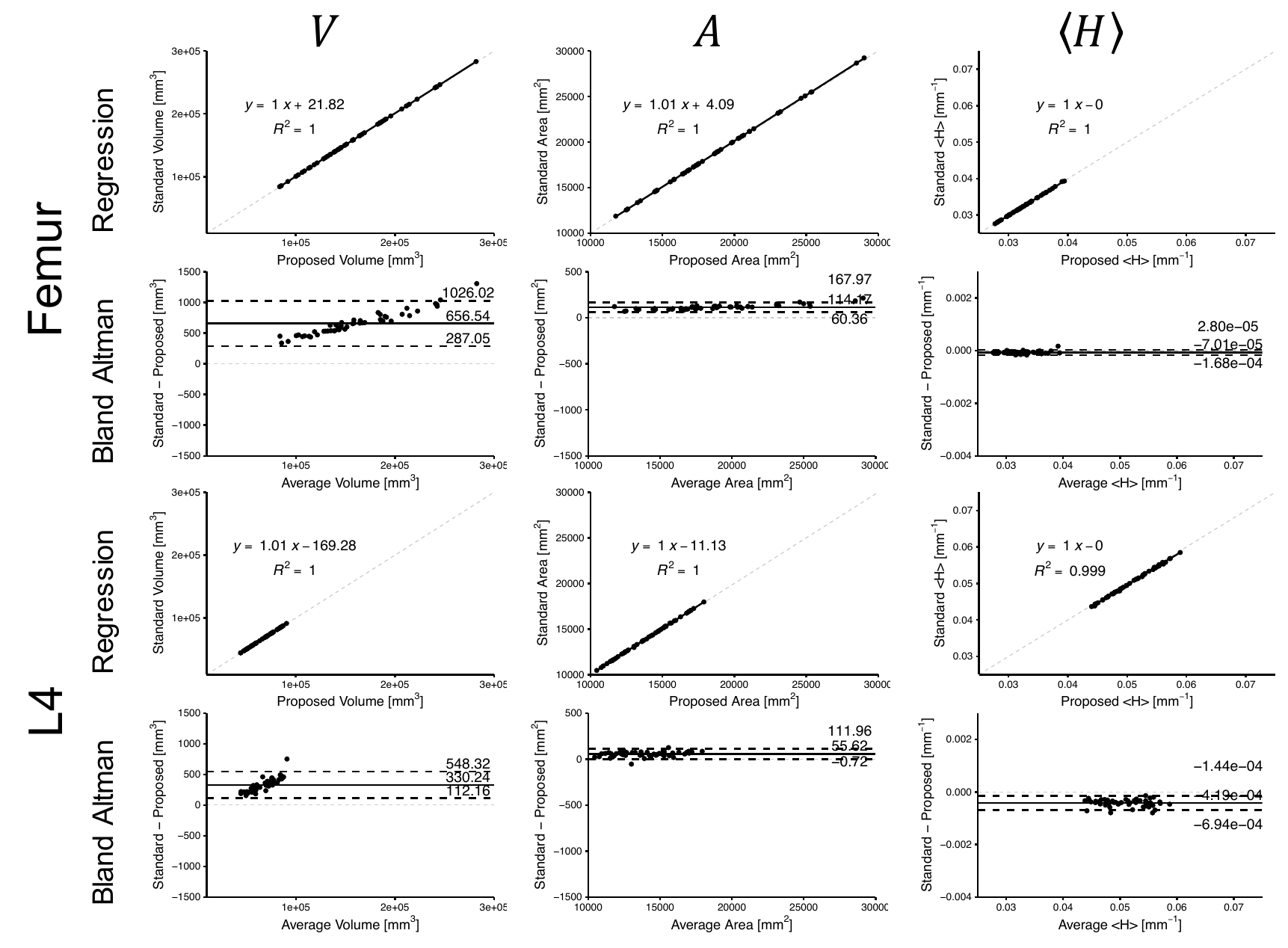}%
    \caption{Regression and Bland-Altman plots demonstrating agreement between the proposed and standard methods for the femur and L4 vertebra (n=50 in each group). Dashed light gray lines denote ideal relationships.}
    \label{fig:regression-ba-proposed}
  \end{figure}
\end{landscape}

Regression and Bland-Altman plots for volume, area, and average mean surface curvature are displayed in Figure~\ref{fig:regression-ba-sdt} for the signed distance transform and Figure~\ref{fig:regression-ba-proposed} for the proposed embedding.
Regression and Bland-Altman statistics for both embeddings are summarized in Table~\ref{tab:agreement}.
The proposed method reduces variability in area and average mean curvature measures, while greatly reducing the proportional bias in area as compared to the signed distance transform.
Computed as the difference in limits of agreement over the average between methods, the area proportional bias slope improved from -5.0\% in the femur and -3.1\% in the vertebrae to 0.6\% in the femur and 0.8\% in the vertebrae using the proposed method.
Regression slopes all improved or remained unity.
Excellent agreement is seen between the proposed and traditional methods.
While the global morphometric outcomes appear reasonable for the signed distance transform, Figure~\ref{fig:sdt} demonstrates that the local measures are quantized and averaging across the surface has increased global outcome accuracy.

\begin{landscape}
  \begin{table}
    \footnotesize
    \centering
    \begin{threeparttable}
    \begin{tabular}{llcllclllcll} 
      & & & \multicolumn{2}{c}{Descriptive Statistics\tnote{$\dagger$}} & & \multicolumn{3}{c}{Regression Analysis} & & \multicolumn{2}{c}{Bland-Altman Analysis} \\
      \cline{4-5}\cline{7-9}\cline{11-12}
      & & & Method & Standard & & Slope (95\% CI) & Intercept (95\% CI) & R\textsuperscript{2} & & Bias (95\% LoA) & p-value\tnote{$\ddagger$}\\
      \hline
      \multicolumn{12}{c}{Signed Distance Transform} \\
      \hline
      \multirow{3}{*}{Femur} & $V~[\si{\milli\metre\cubed}]$ & & $159959.74 \pm 45821.52$ & $160513.90 \pm 45992.36$ & & $1.0037~(1.0034, 1.0040)$ & $-42.12~(-94.53, 10.29)$ & $1.000$ & & $554.16~(205.45, 902.88)$ & $<2\cdot 10^{-16}$ \\
      & $A~[\si{\milli\metre\squared}]$ & & $19669.37 \pm 4213.27$ & $18775.06 \pm 3991.35$ & & $0.9470~(0.9397, 0.9543)$ & $148.27~(1.76, 294.77)$ & $0.999$ & & $-894.31~(-1378.64, -409.97)$ & $<2\cdot 10^{-16}$ \\
      & $\langle H \rangle~[\si{\per\milli\metre}]$ & & $0.0319 \pm 0.0029$ & $0.0324 \pm 0.0029$ & & $0.9615~(0.8900, 1.0329)$ & $0.0018~(-0.0005, 0.0041)$ & $0.938$ & & $0.0006~(-0.0009, 0.0020)$ & $0.834$ \\
      \cline{2-12}
      \multirow{3}{*}{L4} & $V~[\si{\milli\metre\cubed}]$ & & $66762.55 \pm 12878.85$ & $67009.42 \pm 12961.63$ & & $1.0064~(1.0055, 1.0073)$ & $-181.96~(-242.27, -121.65)$ & $1.000$ & & $246.86~(67.29, 426.43)$ & $<2\cdot 10^{-16}$ \\
      & $A~[\si{\milli\metre\squared}]$ & & $14829.09 \pm 2049.11$ & $14110.63 \pm 1971.38$ & & $0.9613~(0.9499, 0.9727)$ & $-144.0273~(-314.79, 26.73)$ & $0.998$ & & $-718.47~(-940.13, -496.81)$ & $3.5\cdot 10^{-8}$ \\
      & $\langle H \rangle~[\si{\per\milli\metre}]$ & & $0.0515 \pm 0.0044$ & $0.0507 \pm 0.0040$ & & $0.9010~(0.0009, 0.0078)$ & $0.0043~(0.0009, 0.0078)$ & $0.939$ & & $-0.0008~(-0.0029, 0.0013)$ & $0.0473$ \\
      \hline
      \multicolumn{12}{c}{Proposed Method} \\
      \hline
      \multirow{3}{*}{Femur} & $V~[\si{\milli\metre\cubed}]$ & & $159857.37 \pm 45810.44$ & $160513.90 \pm 45992.36$ & & $1.0040~(1.0037, 1.0043)$ & $21.82~(-30.31, 73.96)$ & $1.000$ & & $656.54~(287.05, 1026.02)$ & $<2\cdot 10^{-16}$ \\
      & $A~[\si{\milli\metre\squared}]$ & & $18660.90 \pm 3967.92$ & $18775.06 \pm 3991.35$ & & $1.0059~(1.0049, 1.0069)$ & $4.09~(-15.92, 24.11)$ & $1.000$ & & $114.17~(60.36, 167.97)$ & $3.46\cdot 10^{-15}$ \\
      & $\langle H \rangle~[\si{\per\milli\metre}]$ & & $0.0325 \pm 0.0029$ & $0.0324 \pm 0.0029$ & & $1.0024~(0.9975, 1.0074)$ & $-0.0001~(-0.0003, 0.0000)$ & $1.000$ & & $0.0000~(-0.0002, 0.0000)$ & $0.2972$ \\
      \cline{2-12}
      \multirow{3}{*}{L4} & $V~[\si{\milli\metre\cubed}]$ & & $66679.18 \pm 12865.13$ & $67009.42 \pm 12961.63$ & & $1.0075~(1.0062, 1.0087)$ & $-169.28~(-254.42, -84.14)$ & $1.000$ & & $330.24~(112.16, 548.32)$ & $3.78\cdot 10^{-16}$ \\
      & $A~[\si{\milli\metre\squared}]$ & & $14055.01 \pm 1961.87$ & $14110.63 \pm 1971.38$ & & $1.0047~(1.0007, 1.0088)$ & $-11.13~(-68.20, 45.94)$ & $1.000$ & & $55.62~(-0.72, 111.96)$ & $0.019$ \\
      & $\langle H \rangle~[\si{\per\milli\metre}]$ & & $0.0511 \pm 0.0041$ & $0.0507 \pm 0.0040$ & & $0.9958~(0.9858, 1.0057)$ & $-0.0002~(-0.0007, 0.0003)$ & $0.999$ & & $-0.0004~(-0.0007, -0.0001)$ & $0.466$ \\
      \hline
    \end{tabular}
    \begin{tablenotes}
      \item[$\dagger$] Reported mean $\pm$ standard deviation.
      \item[$\ddagger$] Computed on slope in the Bland-Altman diagram to test for proportional bias.
    \end{tablenotes}
    \caption{Agreement analysis between for the signed distance transform and proposed method relative to the standard methods for continuous outcomes.}
    \label{tab:agreement}
    \end{threeparttable}
  \end{table}
\end{landscape}

At the $\alpha = 0.05$ level, statistically significant proportional bias is seen in measures of area and volume using the proposed method.
While the proportional bias is statistically significant, it is practically insignificant exhibiting a slope of less than 1\% for all measures.
The source of the proportional bias is believed to be an interaction between the Gaussian filtration and regularized Heaviside and Dirac delta functions, where there is a sub-voxel shift of no more than one voxel in the embedding with a direction and magnitude that depends on the local mean curvature.
The embedding shrinks at areas of positive mean curvature and expands at areas of negative mean curvature (Figure~\ref{fig:gauss}).
Although locally small, the volume and area integrals accumulate the error across the volume into a detectable bias.
This leads to a proportional bias because the larger the object, the more accumulation.
It should be noted that this is independent from the structural changes caused by Gaussian blurring as the analysis was performed on the data re-binarized after embedding.

\begin{figure}
  \centering
  \includegraphics[width=\linewidth]{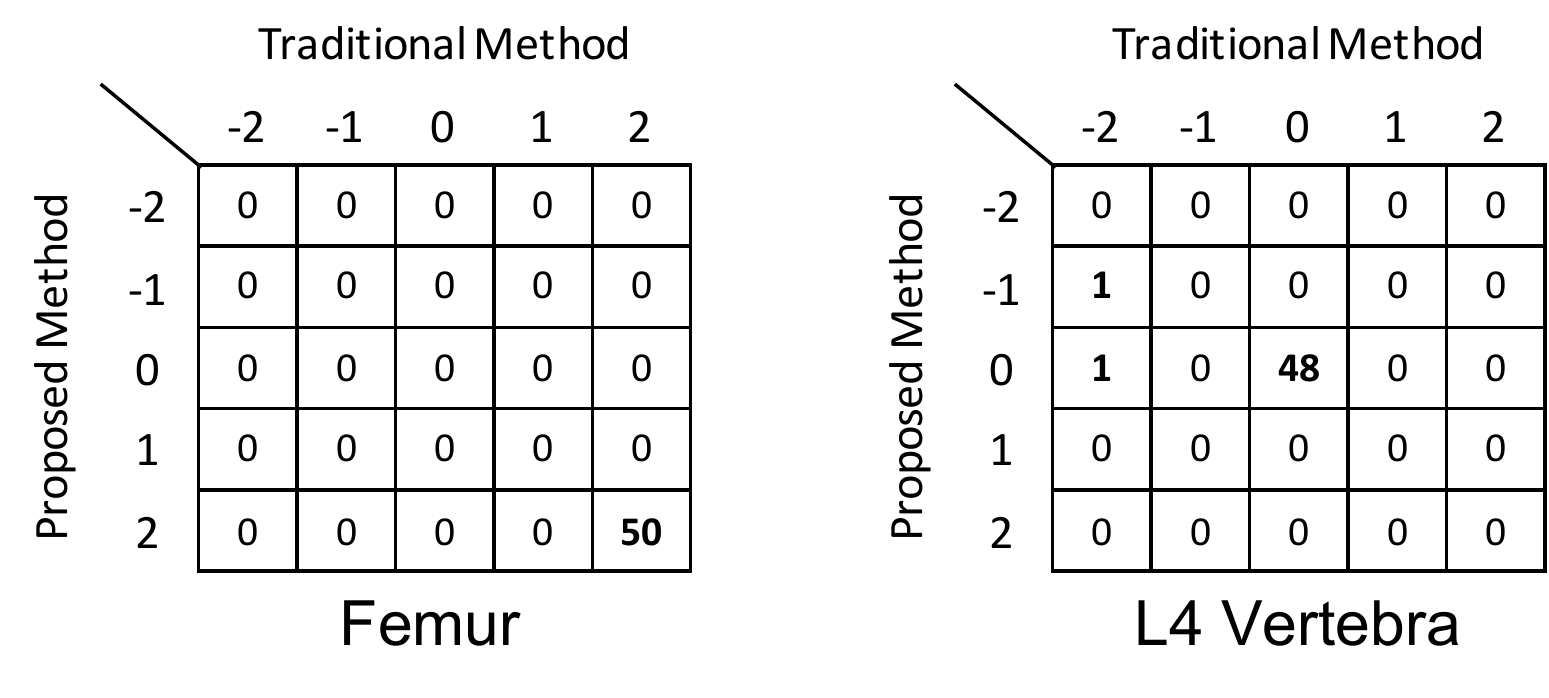}%
  \caption{Confusion matrix demonstrating agreement between the proposed and standard method for computing the Euler-Poincaré characteristic in the femur and L4 vertebra (n=50 in each group).}
  \label{fig:confusion_matrix}
\end{figure}

\begin{figure}[h]
  \centering
  \includegraphics[width=0.8\linewidth]{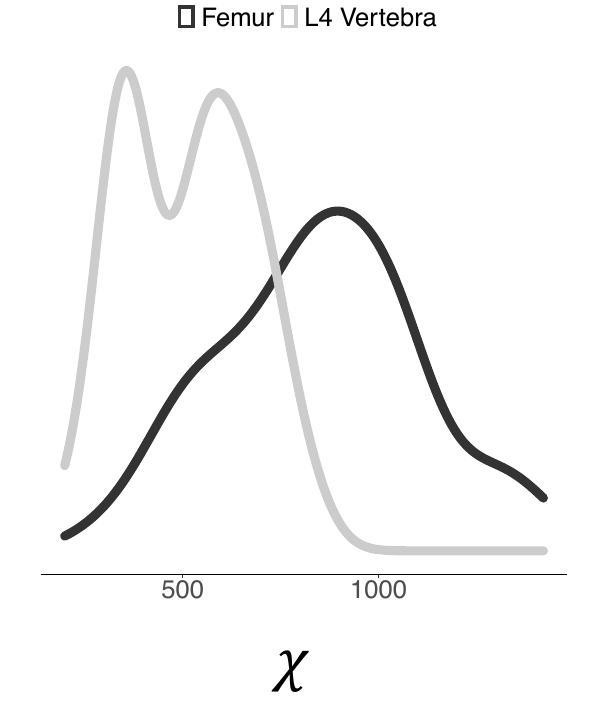}%
  \caption{Histogram of Euler-Poincaré characteristic in the femur and L4 vertebra (n=50 in each group) computed using the signed distance transform embedding. Values should be around 0 and 2 but instead are erroneously larger.}
  \label{fig:chi_histogram}
\end{figure}

Confusion matrices for the computation of the \epc~characteristic are given in Figure~\ref{fig:confusion_matrix}. 100\% accuracy is seen in the femur and 96\% accuracy is seen in the L4 vertebrae.
The \epc~characteristic was nonsensical when computed from the signed distance transform, so the results are visualized using a histogram rather than the confusion matrix (Figure~\ref{fig:chi_histogram}).
The reason for such a large errors is that the quantization in the signed distance transform makes the embedding at best $O(h)$ accurate in initialization~\cite{besler2020artifacts}, causing the second derivatives to amplify noise.
This result is fundamental to the signed distance transform and cannot be corrected by increasing the resolution or transforming the signed distance transform in some way~\cite{besler2020artifacts}.

\begin{figure}
  \centering
  \includegraphics[width=\linewidth]{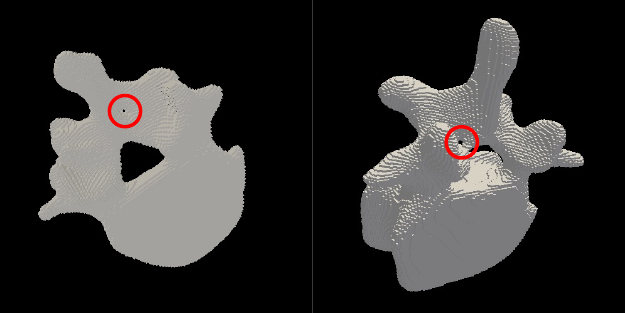}%
  \caption{A second hole in two L4 vertebrae, highlighted in red. This could be due to a contouring error or due to an anatomical defect in the pedicles.}
  \label{fig:l4_hole}
\end{figure}

Two vertebrae display an \epc~characteristic of -2 indicating a second hole in their shape.
These images are displayed in Figure~\ref{fig:l4_hole}.
One vertebra has a hole of a single voxel while another has a hole of a few voxels.
As the regularization thickness is larger than these holes, an error is seen in the proposed method’s computation of \epc~characteristic giving 0 in the first case and -1 in the second.
Most likely, there are more small holes in the surfaces caused by segmentation errors or anatomical defects that are removed with Gaussian filtering.

\begin{figure*}
  \centering
  \includegraphics[width=\linewidth]{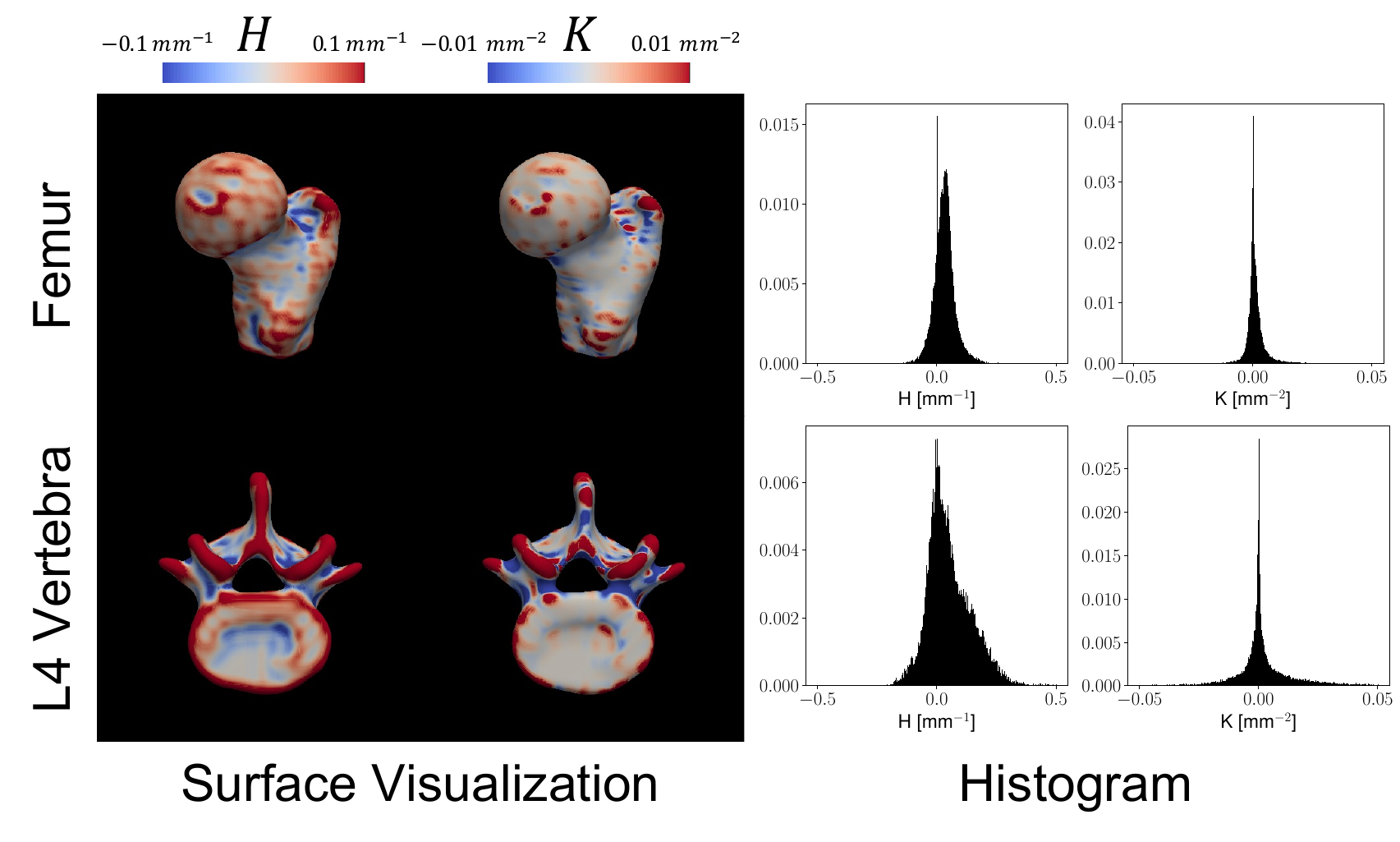}%
  \caption{(Left) Visualization of local mean and Gaussian curvature for a femur and L4 vertebra. (Right) histogram of mean and Gaussian curvature across the surface.}
  \label{fig:surface_overview}
\end{figure*}

A femur and L4 vertebra, selected as the objects with median surface average mean curvature, are rendered in Figure~\ref{fig:surface_overview}.
Local mean curvature is consistent with intuition: the fovea capitis has a negative mean curvature and the tips of the transverse processes show a positive mean curvature.
The mean and Gaussian histograms are fat-tailed distributions with the L4 mean curvature histogram exhibiting skewness.
The femur mean curvature histogram has a large spike at zero corresponding to the flat distal section where the scan volume of interest cropped the femur.

\section{Discussion}
The main contribution of this paper is the description of a method for performing morphometry on closed, implicit surfaces.
For this purpose, an embedding procedure for binary images based on Gaussian blurring is suggested as an alternative to the signed distance transform to avoid errors from quantization.
Morphometrics resulting from this embedding are validated against well-established methods and show excellent agreement and considerably better results compared to using the standard signed distance transform.

The proposed method is a refinement and summation of many classic works~\cite{sethian1999level}.
Measuring area and volume from implicit surfaces is well-defined~\cite{chan2001active} while total Gaussian curvature has been previously used to count the number of objects in a volume preserving flow~\cite{peng1999pde}.
Mean curvature has been used extensively in the computation of mean curvature flow~\cite{osher1988fronts,chopp1991computing}.
The main contribution of this work is a well-defined embedding function and synthesis of previously described methods into morphometrics founded in differential geometry.
This technique provides a way to measure mathematically well-described properties on anatomical structures for basic or clinical research.

The embedding method based on Gaussian blurring is the key to enable an advanced morphometric analysis since computation of curvatures from signed distance transforms of binary images result in considerable quantization-related errors~\cite{besler2020artifacts}.
These errors can have a particular negative effect on the computation of \epc~characteristic and will be most obvious and severe in that outcome first.
In essence, the problem is to assign well defined spatial gradients to binary images.

The consequences of the quantized embedding are more profound than just morphometry.
An error in the representation of embeddings can also lead to irreducible error in curve evolution problems~\cite{coquerelle2016fourth} that are independent of voxel spacing~\cite{besler2020artifacts}.
This highly motivates the use of flexible, local level set initialization methods~\cite{li2005level} for active contour problems.

Embedding methods can be designed for specific applications but generally require that the gradients are accurate.
We previously proposed a dithering and reinitialization algorithm~\cite{peng1999pde} for fixing the gradient issue in signed distance transforms.
While the algorithm improves accuracy compared to using signed distance transforms directly, the method does not produce the accuracy seen in this work since the algorithm stops improving before gradients of the embedding are highly accurate (data not shown).
Furthermore, the algorithm requires an exceptional amount of computation time limiting practical application.
The proposed method was inspired by a sub-voxel distance mapping method~\cite{caselles1993geometric} and flexible, local level set initialization~\cite{li2005level}.
The point remains that there is space for design around the embedding method.
It is important to highlight that in some workflows, no embedding procedure is needed as the data already comes embedded.
This is true in active contour segmentation models~\cite{chan2001active,vese2002multiphase,cremers2007review} in particular.

The general design criterion for embeddings is that the Heaviside recovers the object (Equation~\ref{eqn:recovery} and~\ref{eqn:level_set}) and that the embedding is monotonic across the zero level set.
Practically, the embedding only needs to be defined near the surface --- the so-called narrow band method~\cite{adalsteinsson1995fast}.
A Gaussian filter was used due to its speed, ease of design, and widespread applicability.
However, other methods such as anisotropic diffusion~\cite{perona1990scale}, anti-aliasing filters~\cite{whitaker2000reducing}, or simple non-nearest neighbor interpolation~\cite{thevenaz2000image} could also be used and may prove beneficial.
The design objective is that 1) the Heaviside recovers the object, 2) gradients are accurate, and 3) features relevant for experimental work are not removed.
Within this context, embedding with a Gaussian filter has advantages and disadvantages.
The major disadvantage is that Gaussian filter modifies the underlying object.
Objects thinner or closer together than the full-width half maximum of the blurring Gaussian are likely to be closed or opened.
Nevertheless, large organs relative to image spacing such as long bones, the hippocampus, and the liver are unlikely to change.
One advantage is that Gaussian filtration also helps to remove small imperfections in the binary images (e.g. manual segmentation artifacts).
In this work, manual contouring artifacts were seen in the data, which would have appeared as noise in the curvature outcomes.
The Gaussian standard deviation is an intuitive and obvious parameter to handle this artifact.
Furthermore, it helps handling images of varying resolutions.
As the image resolution increases, smaller dimples in the surface can be resolved increasing the absolute value of the curvature that can be measured.
By filtering the data at a physical size, these small differences between datasets can be standardized.

The primary advantage of the proposed method is that it is local.
In this context, local means that the morphometrics can be evaluated locally in the image while global means the values can only be computed for the surface as a whole.
Global methods exist for the computation of the mean curvature~\cite{hildebrand1997quantification,hahn1992trabecular,jinnai2002surface} and \epc~characteristic~\cite{odgaard1993quantification} while binarizing the embedding allows computation of volume and area. Local curvature can also be evaluated from a mesh of the surface \cite{goldfeather2004novel,rusinkiewicz2004estimating,flynn1989reliable}.
However, meshes are not ideal for curve evolution problems due to the splitting and merging required to change topology.
Having the ability to evaluate these outcomes locally opens up many possibilities.
First, they can be visualized and correlated with other measures such as local stress from finite element analysis~\cite{loundagin2020stressed} and local bone formation rates~\cite{schulte2011vivo}.
Second, they can be used as a loss function in deep learning models~\cite{litjens2017survey} because gradients in back propagation can be defined through the spatial gradients.

An important feature of the proposed method is that the computation of \epc~characteristic is not limited to integer values because a continuous measure is integrated across the surface.
As was seen in the vertebra, this can produce misleading results when small holes relative to the regularization thickness are present in the structure.
This will not be an issue in many cases and the \epc~characteristic can be rounded to an integer.
However, this mistake will be obvious to spot in the resulting data if odd values or values greater than 2 are measured for the \epc~characteristics.

The main limitation of this study is that local curvatures were not directly validated as we assumed that if the global morphometry is accurate, the local morphometry will also be accurate.
However, this may not be the case since averaging across the surface should increase the accuracy of the results.
Given that the \epc~characteristic is not averaged and accumulates errors across the surface, it is reasonable to assume the error in local curvature is small.
Additionally, qualitative analyses based on visualizing curvatures on the surface provided evidence that local measures are accurate.
In the future, ideal parametric surfaces such as spheres, tori, or triply period minimal surfaces~\cite{schoen1970infinite} where curvatures can be computed analytically should be used for validation.

\section{Conclusion}
A method of computing volume, area, average mean curvature, and \epc~characteristic of closed, orientable surfaces is described.
The fast and simple Gaussian fitler is proposed for embedding binary images to overcome the quantization errors associated with the signed distance transform. The method is accurate and local, allowing the visualization of curvatures across the surface.

\appendix
\section{Regularization Selection}
\label{app:regulariz_selection}
The regularization parameter $\epsilon$ is easy to select for the signed distance transform since the embedding encodes physical dimensions.
With the embedding of Equation~\ref{eqn:embedding}, the embedding no longer relates to physical sizes.
The objective of regularization selection is to select $\epsilon$ such that area integrals average over the same physical thickness in the image domain.

Let $t$ denote the physical thickness over which to integrate.
The objective is to select $\epsilon$ such that the response of the Dirac delta function of the embedding has support $t=x_2-x_1$.
\begin{eqnarray}
  \lvert \phi \rvert < \epsilon
\end{eqnarray}
Since $\phi$ is monotonically decreasing and symmetric around the zero crossing, this can be equivalently stated as:
\begin{equation}
  \label{eqn:epsilon_phi}
  2\epsilon  = \phi(x_1) - \phi(x_2)
\end{equation}
The problem is now to find an expression for the embedding.
Reiterating Equation~\ref{eqn:embedding}, the embedding equation is known:
\begin{equation}
  \phi = T - G_\sigma * I
\end{equation}
The n-dimensional problem can be reduced to one dimension by aligning along the normal of the object and treating it locally as a Heaviside function with the object edge located at $x=0$.
\begin{eqnarray}
  \phi(x) = T - G_\sigma * \theta(x)
\end{eqnarray}
The convolution operator can be evaluated giving an intuitive equation for the filter response:
\begin{equation}
  \label{eqn:edge}
  \phi(x) = T - \frac{1}{2} \erf\left(\frac{x}{\sqrt{2}\sigma}\right)
\end{equation}
where $\erf(\cdot)$ is the error function common in statistics:
\begin{equation}
  \erf(z) = \frac{2}{\sqrt{\pi}} \int_0^z e^{-t^2} dt
\end{equation}
Finally, the regularization parameter can be solved for by substituting Equation~\ref{eqn:edge} into Equation~\ref{eqn:epsilon_phi} noting that $x_2=t/2$ and $x_1=-t/2$:
\begin{equation}
  \epsilon = \frac{1}{2} \erf\left(\frac{t}{2\sqrt{2}\sigma}\right)
\end{equation}
To gain intuition, the embedding equation can be linearized around the point $x=0$.
\begin{eqnarray}
  \phi(x) & \approx & \phi(0) + \phi^\prime(0)(x-0) \\
  \phi(x) & \approx & (T - \frac{1}{2}) - \frac{x}{\sqrt{2\pi}\sigma}
\end{eqnarray}
Substituting into Equation~\ref{eqn:epsilon_phi},
\begin{eqnarray}
  & 2\epsilon = \left[T - \frac{1}{2} - \frac{x_1}{\sqrt{2\pi}\sigma}\right] - \left[T - \frac{1}{2} - \frac{x_2}{\sqrt{2\pi}\sigma}\right] \\
  \label{eqn:linear}
  & \epsilon =  \frac{t}{2 \sigma \sqrt{2 \pi}} \approx \frac{5t}{6}
\end{eqnarray}
Equation~\ref{eqn:linear} differs from the full-width half-maximum response of a Gaussian by a factor of $\sqrt{\pi / \ln2}$.
The linearization is good for $t \ll \sigma$.
The regularization thickness must be selected less than the filter support, which is a few multiples of $\sigma$.
The regularization is a function of standard deviation only and does not vary with the threshold.

\section{Finite Difference Stencils}
\label{app:stencils}
For posterity, the exact finite difference equations are given.
The notation $\phi_i = \phi(x+ih_x)$ and $\phi_{i,j} = \phi(x+ih_x,y+jh_y)$ is used to keep equations brief.
Finite differences are defined along the x and y directions but are equivalent in all directions.
The standard and well-defined first, second, and mixed finite differences are as such: 
\begin{eqnarray}
  \label{eqn:fd:x}
  \phi_x \approx D^x\phi = \frac{
    \frac{1}{12}\phi_{-2} - \frac{2}{3}\phi_{-1} + \frac{2}{3}\phi_{1} - \frac{1}{12}\phi_{2}
    }{
      h_x
    } \\
  \label{eqn:fd:xx}
  \phi_{xx} \approx D^{xx} \phi = \frac{
    -\frac{1}{12}\phi_{-2} + \frac{4}{3}\phi_{-1} - \frac{5}{2}\phi_{0} + \frac{4}{3}\phi_{1} - \frac{1}{12}\phi_{2}
    }{
      h_x^2
    } \\
  \label{eqn:fd:xy}
    \begin{aligned}
      \phi_{xy} \approx D^{xy}\phi & = \frac{1}{144 h_x h_y} \cdot \\
      & \left[ 8 \left(\phi_{1,-2} + \phi_{2,-1} + \phi_{-2,1} + \phi_{-1,2}\right) \right. \\
      & \left. - 8 \left(\phi_{-1,-2} + \phi_{-2,-1} + \phi_{2,1} + \phi_{1,2}\right) \right. \\
      & \left.  - \left(\phi_{2,-2} + \phi_{-2,2} - \phi_{-2,-2} - \phi_{2,2}\right) \right. \\
      & \left. + 64 \left(\phi_{-1,-1} + \phi_{1,1} - \phi_{1,-1} - \phi_{-1,1}\right) \right]
    \end{aligned}
\end{eqnarray}
Additionally, the equations for mean and Gaussian curvature are expanded for completeness~\cite{sethian1999level}:
\begin{eqnarray}
  \label{eqn:mean_curvature_numerical}
  H(\phi) = \frac{
    \begin{aligned}
      \left(\phi_{yy} + \phi_{zz}\right)\phi_x^2
    + \left(\phi_{zz} + \phi_{xx}\right)\phi_y^2 \\
    + \left(\phi_{xx} + \phi_{yy}\right)\phi_z^2 \\
    -2 \phi_x\phi_y\phi_{xy}
    -2 \phi_z\phi_x\phi_{zx}
    -2 \phi_y\phi_z\phi_{yz}
  \end{aligned}
  }{
  \left(\phi_x^2 + \phi_y^2 + \phi_z^2\right)^\frac{3}{2}
  } \\
  \label{eqn:gauss_numerical}
  K(\phi) = \frac{
    \begin{aligned}
      \phi_x^2\left(\phi_{yy}\phi_{zz} - \phi_{yz}^2\right) \\
    + \phi_y^2\left(\phi_{xx}\phi_{zz} - \phi_{xz}^2\right) \\
    + \phi_z^2\left(\phi_{xx}\phi_{yy} - \phi_{xy}^2\right) \\
    +2 \phi_x\phi_y\left(\phi_{xz}\phi_{yz} - \phi_{xy}\phi_{zz}\right) \\
    +2 \phi_y\phi_z\left(\phi_{xy}\phi_{xz} - \phi_{yz}\phi_{xx}\right) \\
    +2 \phi_x\phi_y\left(\phi_{xy}\phi_{yz} - \phi_{xz}\phi_{yy}\right) \\
    \end{aligned}
  }{
    \left(\phi_x^2 + \phi_y^2 + \phi_z^2\right)^2
  }
\end{eqnarray}

Finite differences are substituted directly into their respective location in Equations~\ref{eqn:mean_curvature_numerical} and ~\ref{eqn:gauss_numerical}.



\bibliographystyle{cas-model2-names}

\bibliography{Binary-Morph}





\end{document}